\documentclass[
]{ceurart}

\usepackage{xcolor}
\usepackage{booktabs}

\usepackage{bbold}
\usepackage{multirow}
\usepackage{array}
\usepackage{amsmath}
\usepackage{hyperref}
\usepackage[table]{xcolor}
\usepackage[most]{tcolorbox}
\usepackage{cellspace, makecell, multirow}
\usepackage{amsthm}
\usepackage{threeparttable}

\usepackage{listings}
\begin{document}

\copyrightyear{2025}
\copyrightclause{Copyright for this paper by its authors.
  Use permitted under Creative Commons License Attribution 4.0 International (CC BY 4.0).}

\conference{ITADATA2025: The 4$^{\text{th}}$ Italian Conference on Big Data and Data Science, September 9--11, 2025, Turin, Italy}

\title{DART: A Structured Dataset of Regulatory Drug Documents in Italian for Clinical NLP}

\author[1,2]{Mariano Barone}
[%
orcid=0009-0004-0744-2386,
email=mariano.barone@unina.it,
url=https://github.com/csmariano,
]

\author[1]{Antonio Laudante}
[%
orcid=,
email=antonio.laudante@unina.it,
url=,
]

\author[1,2]{Giuseppe Riccio}
[%
orcid=0009-0002-8613-1126,
email=giuseppe.riccio3@unina.it,
url=https://github.com/giuseppericcio,
]
\cormark[1]

\author[1,2]{Antonio Romano}
[%
orcid=0009-0000-5377-5051,
email=antonio.romano5@unina.it,
url=https://github.com/LaErre9,
]

\author[3]{Marco Postiglione}
[%
orcid=0000-0001-6092-940X,
email=marco.postiglione@northwestern.edu,
]

\author[1,2]{Vincenzo Moscato}
[%
orcid=0000-0002-0754-7696,
email=vincenzo.moscato@unina.it,
url=http://wpage.unina.it/vmoscato/,
]

\address[1]{University of Naples Federico II, Department of Electrical Engineering and Information Technology (DIETI), Via Claudio, 21 - 80125 - Naples, Italy}
\address[2]{Consorzio Interuniversitario Nazionale per l'Informatica (CINI) - ITEM National Lab, Complesso Universitario Monte S.Angelo, Naples, Italy}
\address[3]{Northwestern University, Department of Computer Science, McCormick School of Engineering and Applied Science, 2233 Tech Dr, Evanston, IL 60208, United States}

\cortext[1]{Corresponding author.}

\begin{abstract}
The extraction of pharmacological knowledge from regulatory documents has become a key focus in biomedical natural language processing, with applications ranging from adverse event monitoring to AI-assisted clinical decision support. However, research in this field has predominantly relied on English-language corpora such as DrugBank, leaving a significant gap in resources tailored to other healthcare systems. To address this limitation, we introduce \texttt{DART} (Drug Annotation from Regulatory Texts), the first structured corpus of Italian Summaries of Product Characteristics derived from the official repository of the Italian Medicines Agency (AIFA). The dataset was built through a reproducible pipeline encompassing web-scale document retrieval, semantic segmentation of regulatory sections, and clinical summarization using a few-shot-tuned large language model with low-temperature decoding. \texttt{DART} provides structured information on key pharmacological domains such as indications, adverse drug reactions, and drug–drug interactions. To validate its utility, we implemented an LLM-based drug interaction checker that leverages the dataset to infer clinically meaningful interactions. Experimental results show that instruction-tuned LLMs can accurately infer potential interactions and their clinical implications when grounded in the structured textual fields of \texttt{DART}. We publicly release our code on GitHub: \url{https://github.com/PRAISELab-PicusLab/DART}.
\end{abstract}


\maketitle

\section{Introduction}

In recent times, extracting and organizing pharmacological information from regulatory documents has taken on a pivotal role in the domain of biomedical natural language processing (NLP). This research goal is focused on automating the assimilation of clinical and regulatory data into decision-oriented processes \cite{velupillai2018using}, thereby supporting applications like prescription aid systems and pharmacovigilance instruments. Among these regulatory resources, the Summary of Product Characteristics (SmPC) --- referred to in Italy as the Riassunto delle Caratteristiche del Prodotto (RPC) --- is notably distinguished as a comprehensive and reliable document published by the Italian Medicines Agency\footnote{\url{https://www.aifa.gov.it/}} (AIFA). Designed for healthcare professionals, the RCP serves as the 'identity card' of a medicinal product, providing standardized and regularly updated information on efficacy, safety, therapeutic use, contraindications, adverse drug reactions (ADR), drug-drug interactions (DDI), and other essential clinical characteristics.
Despite its importance, such texts remain underrepresented in the literature, with most prior work focusing exclusively on English-language corpora and overlooking the linguistic and structural particularities of national regulatory frameworks. In the Italian context, the absence of tailored resources hampers the development of clinically grounded AI (Artificial Intelligence) systems that align with local healthcare practices and regulatory standards. To address this gap, we present \texttt{DART} (Drug Annotation from Regulatory Texts), a structured corpus of RCPs in Italian, developed through a scalable and reproducible pipeline. The dataset is built by automatically retrieving documents from AIFA, extracting and semantically segmenting their contents, and organizing the information into structured fields that correspond to standard regulatory sections. Additionally, \texttt{DART} is enhanced with clinical summaries generated using large language models (LLMs) via few-shot learning and low-temperature decoding strategies. These summaries are intended to support downstream applications such as interaction checking, knowledge graph construction, and automated risk profiling.
With more than 16,000 processed RCPs and over 95 million tokens, \texttt{DART} represents a high-value asset for the Italian clinical NLP community and the broader healthcare data science ecosystem. It provides a robust foundation for the training, evaluation, and deployment of large-scale language models in both regulatory and clinical contexts. Furthermore, \texttt{DART} contributes significantly to the healthcare Big Data ecosystem by offering a high-resolution corpus of regulatory texts that supports the training of LLMs, the development of interpretable knowledge graphs, and the implementation of AI-driven clinical decision-making tools.

\section{Related Work}

The extraction of pharmacological knowledge from regulatory texts—such as Summary of Product Characteristics (SmPC) —is a growing area in biomedical NLP. These documents offer authoritative information on adverse drug reactions (ADRs), drug–drug interactions (DDIs), contraindications, and indications, and form the normative basis for safe prescribing. However, most existing work has focused on English-language corpora, leaving national regulatory texts, especially Italian RCPs, underrepresented. Early ADR extraction relied on classical machine learning models, including ensemble methods and multilayer perceptrons~\cite{hong2023recent, DBLP:journals/access/AbbasSAAKZ23}. The adoption of transformer-based architectures such as BERT, BioBERT, and PubMedBERT significantly improved performance~\cite{DBLP:conf/eacl/PortelliLCSS21}, though non-English texts still require costly adaptation and fine-tuning~\cite{DBLP:conf/clef/RomanoRPM24}. More recently, large language models (LLMs) like GPT-4 have shown strong zero- and few-shot performance in biomedical tasks, including ADR detection, outperforming traditional baselines and enhancing interpretability in pharmacovigilance pipelines~\cite{llmShortADR2025, DBLP:conf/atsip/AlshehriKSA24, DBLP:conf/fllm/AbuNasserAJDA24}. Retrieval-augmented generation and agent-based simulation approaches have further demonstrated the value of context-aware models~\cite{DBLP:conf/bigdataconf/RussoRORRGPM24, DBLP:conf/asunam/FerraroGGPORRRM24}. DDI prediction has similarly evolved toward hybrid and graph-based architectures. Recent studies integrate knowledge graphs (KGs) with LLMs to produce accurate and explainable predictions~\cite{xu2024ddi, DBLP:conf/bigdataconf/RussoORRGPM24}, while in-context learning techniques have improved interaction detection~\cite{qi2025improving}. Medication recommendation systems now incorporate regulatory text and clinical narratives, outperforming structured-code-based methods, especially in multilingual and safety-aware settings~\cite{llmDrugRecommendation2024, kim2024medication, DBLP:conf/ecir/RomanoRPM25}. Ongoing work also explores explainability in recommendations~\cite{ferraro2024explanation} and the combination of symbolic and generative approaches in medical summarization~\cite{DBLP:conf/itadata/RiccioRKCPGFGM23}. Despite this progress, Italian regulatory documents remain largely unexplored. Resources like DrugBank~\cite{drugbank2024} include Italian drug names but abstract away regulatory phrasing and section structure. Challenges such as DIMMI~\cite{dimmi2024} and aggregation efforts~\cite{DBLP:journals/jbi/AyvazHHZSTVBSRD15} highlight the need for domain-specific resources aligned with the Italian context. Real-world data sources like the National Pharmacovigilance Network (RNF)~\cite{ruggiero2020immune, magro2021identifying}, VALORE~\cite{trifiro2021large}, and regional datasets~\cite{galai2022time} offer complementary insights but are often incomplete, misaligned with regulatory language, or dependent on patient self-reporting~\cite{leone2010identifying}. In contrast, RCPs provide standardized, high-quality knowledge that enables direct modeling of pharmacological phenomena~\cite{shen2021automatic}. To fill this gap, we present \texttt{DART}, a structured dataset derived from full-text Italian RCPs, designed to support the development of LLM-based systems grounded in official regulatory content. DART enables validation of LLM outputs against both normative sources and observational datasets, fostering a bidirectional loop between automated pharmacological reasoning and real-world clinical safety evidence.

\section{Materials and Methods}

\subsection{Dataset Construction}

\begin{figure}[ht]
\centering
\includegraphics[width=0.9\linewidth]{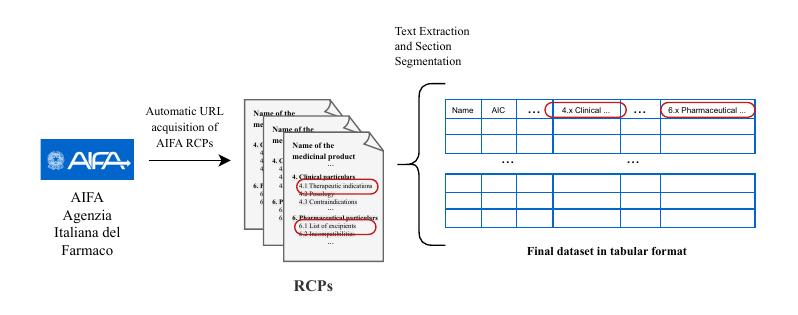}
\caption{End-to-end pipeline for the construction of the \texttt{DART} dataset. The diagram outlines the three main components of the workflow: automated retrieval of RCP documents from AIFA, text extraction and segmentation, and the final structuring and validation of the resulting dataset.}
\label{fig:pipeline}
\end{figure}

The dataset \texttt{DART} was constructed through a three-step pipeline designed for reproducibility and scalability. Specifically: (i) automated retrieval of URLs for the Summary of Product Characteristics (RCPs) from the AIFA portal, (ii) semantic parsing and segmentation of the extracted RCP text, and (iii) data structuring, filtering, and validation. All modules were implemented in Python using open-source libraries. The complete construction workflow is illustrated in Figure~\ref{fig:pipeline}.

\subsubsection{Automated Retrieval of RCP URLs}

The first phase of the pipeline involved the programmatic retrieval of RCP PDFs by interrogating undocumented but publicly accessible RESTful APIs exposed by the AIFA web portal. Due to the SPA-based (Single Page Application) architecture of the website—built using frameworks such as Angular—static DOM scraping was ineffective. Instead, a detailed network traffic analysis was conducted via browser developer tools (\textit{DevTools}, “Network” tab), which led to the identification of two critical endpoints. The first is a search endpoint, which requires a zero-padded AIC code (e.g., \texttt{123456} becomes \texttt{00123456}) and returns a JSON payload containing metadata for each drug, including the keys \texttt{CodiceSis} and \texttt{aic6}. These values are then used to query a second endpoint that provides a direct URL to the corresponding RCP PDF. This two-step API interaction is illustrated in Figure~\ref{fig:aifa_api_example}. Although these APIs are unofficial and subject to change without notice, they provided the only viable and scalable access to RCP documents at the time of this study (June 2025). A web spider was implemented in Python using the \texttt{requests} library and seeded with a list of valid AIC codes, sourced from public datasets or inferred from known numerical intervals, in compliance with applicable ethical and legal constraints. For each AIC code, the system executed: querying the search endpoint, parsing the JSON response, constructing the PDF URL, and downloading the file. Failures and exceptions were handled using structured logging.

\begin{figure}[!t]
\centering
\begin{tcolorbox}[width=0.96\textwidth,
                  colback=gray!5,
                  colframe=blue!20,
                  coltitle=black,
                  title=\textbf{Example API Call for RCP PDF Retrieval},
                  enhanced]

\textbf{Step 1: Query the Search Endpoint}\\
\texttt{\href{https://api.aifa.gov.it/aifa-bdf-eif-be/1.0.0/formadosaggio/ricerca?query=AIC_code&spellingCorrection=true&page=0}{https://api.aifa.gov.it/aifa-bdf-eif-be/1.0.0/formadosaggio/ricerca? query=\{AIC\_code\}\&spellingCorrection=true\&page=0}}

\scriptsize
\smallskip
\emph{Note:} The input \texttt{\{AIC\_code\}} must be a zero-padded version of the original AIC code (e.g., \texttt{123456} $\rightarrow$ \texttt{00123456}).

\small
\smallskip
\textbf{Response:} JSON object containing:
\begin{itemize}
  \item \texttt{CodiceSis} (e.g., \texttt{10004290})
  \item \texttt{aic6} (e.g., \texttt{123456})
\end{itemize}

\medskip
\textbf{Step 2: Construct the PDF Download URL}\\
\texttt{\href{https://api.aifa.gov.it/aifa-bdf-eif-be/1.0.0/organizzazione/CodiceSis/farmaci/aic6/stampati?ts=RCP}{https://api.aifa.gov.it/aifa-bdf-eif-be/1.0.0/organizzazione/\{CodiceSis\}/ farmaci/\{aic6\}/stampati?ts=RCP}}

\scriptsize
\smallskip
\emph{Example:}\\
\texttt{https://api.aifa.gov.it/.../organizzazione/10004290/farmaci/123456/stampati?ts=RCP}

\small
\smallskip
\textbf{Output:} Direct link to the corresponding RCP PDF.
\end{tcolorbox}
\caption{Two-step RESTful API interaction for programmatic retrieval of RCP PDFs from AIFA's infrastructure.}
\label{fig:aifa_api_example}
\end{figure}

\subsubsection{Text Extraction and Section Segmentation}

Once the documents were collected, the pipeline proceeded with text extraction and semantic segmentation. Text was extracted using the \texttt{PyMuPDF} library, selected for its robustness in handling complex layouts, preserving reading order, and maintaining basic spatial formatting where feasible. This approach proved effective in most cases, except for PDFs consisting solely of rasterized images (i.e., scanned documents), which lack an embedded text layer. Approximately 4.1\% of collected PDFs were excluded due to the absence of an embedded text layer, making them incompatible with text-based parsing. These cases are flagged for future integration through OCR modules, which are currently under development. The text structuring phase was based on the automatic identification of section headers, which follow well-defined regulatory conventions in RCPs (e.g., "04.1 Therapeutic Indications", "04.8 Undesirable Effects"). A robust regular expression was designed to recognize both the numerical and textual components of the headers, accounting for typographic variability (e.g., spacing, punctuation, capitalization). This enabled segmentation of each document into blocks corresponding to individual sections, each assigned a standardized label. Sections not detected were marked as "N/A" in the resulting dataset, preserving the structural consistency of the data model.

\paragraph{Tabular Content Handling}

Special attention was devoted to Section 04.8 ("Undesirable Effects") often includes tabular structures. While full table parsing was out of scope in this version, raw text within tables was preserved using \texttt{PyMuPDF}’s line-by-line reading mechanism, which retains spatial alignment. Although columnar relationships are not explicitly modeled, the output allows partial semantic interpretation. Future iterations will integrate table extraction tools such as \texttt{pdfplumber}, \texttt{camelot}, or layout-aware parsing models.

\subsubsection{Data Structuring and Validation}

Extracted data were finally mapped into a tabular dataset, where each row corresponds to an RCP document and each column to a specific regulatory section. Final validation included completeness checks (e.g., verifying the presence of expected sections), spot comparisons between raw documents and extracted text, and analysis of error logs produced by the spider. The combined application of these methods enabled the construction of a coherent, scalable dataset suitable for downstream analyses in pharmacological, linguistic, and computational research contexts. Validation steps included logging and error tracking using \texttt{loguru} and structured output reports. On a random sample of 300 documents, over 97\% of expected sections were correctly identified and segmented. Remaining errors were mostly due to non-standard formatting or OCR failures.

\subsection{Preprocessing and Filtering}

Following initial structuring, \texttt{DART} contains 21,502 drugs, subject to a preprocessing step was applied to improve the consistency and correctness of the dataset. This phase involved regex-based cleaning to standardize formatting, eliminating excess whitespace, resolving punctuation inconsistencies, and uniforming typographic variances in section headers. Documents with empty or flawed text were identified and excluded. The "05.0 Pharmacological Properties" section was removed entirely due to a high rate of missing or unusable data, impacting content density and usability. This led to a dataset with strong structural consistency and semantic integrity, appropriate for various clinical NLP applications. Ultimately, 16,029 documents (74.55\%) were correctly segmented into at least 5 mandatory regulatory sections, while the remaining 25.45\% were removed due to structural problems or incomplete content, often originating from OCR errors or missing data.

\subsection{RCP Summarization} \label{sec: RCP_summarization}

To improve the usability of the dataset for both human users and NLP systems, a summarization phase was introduced to condense long and heterogeneous regulatory sections into standardized clinical summaries. This step facilitates tasks such as text classification, knowledge extraction, semantic search, and decision support, while also enabling rapid inspection by clinicians and analysts. Summaries were generated using \texttt{LLaMA~3.1–405B}\cite{DBLP:journals/corr/abs-2407-21783} through Nvidia NIM API\footnote{\url{https://build.nvidia.com/}}, a state-of-the-art large language model, with a low-temperature setting (0.2) to ensure high consistency and minimal hallucination. Each summary was limited to 450 words and aimed to capture key information on drug interactions, adverse events, contraindications, warnings, and pregnancy-related considerations. To guide the generation, we employed a structured prompt combined with a few-shot learning strategy. Handcrafted examples were prepended to the prompt to ensure alignment with regulatory tone, content structure, and domain terminology. Input text was extracted from seven RCP sections (04.1, 04.2, 04.3, 04.4, 04.5, 04.6, and 04.8) and dynamically inserted into the prompt. The resulting summaries were integrated into the dataset as an additional field, enhancing its value for downstream applications and enabling comparisons with real-world DDI/ADR evidence. A manual review of 100 generated summaries showed a high degree of factual consistency (95\%) and minimal hallucination. Most deviations involved stylistic variation or omission of low-priority details. An expert-based validation protocol is currently under development.

\subsection{Dataset Analysis}

\texttt{DART} consists of 16,029 documents, spanning multiple therapeutic areas and regulatory reimbursement classes. The dataset was last updated in May 2025. The corpus comprises over 95 million tokens, with a vocabulary of 102,749 unique terms. Document lengths are generally compact: the mean token count per document is 177.5 (median: 168.3), with a maximum of 9,512 tokens.

\begin{table*}[ht]
    \centering
    \begin{minipage}{0.49\textwidth}
    \centering
    \caption{Overview statistics of the \texttt{DART} dataset}
    \label{tab:dataset_overview}
    \scalebox{0.95}{
    \rowcolors{2}{white}{gray!10}
    \begin{tabular}{lr}
        \toprule
        \textbf{Statistic} & \textbf{Value} \\ 
        \midrule
        \# Number of Medicines &  16,029\\ 
        \# Number of Therapeutic Classes & 6 \\ 
        Last Update & May 2025 \\ 
        \midrule
        Total Tokens &  95,760,718\\ 
        Unique Vocabulary & 102,749\\ 
        Avg. Tokens per Document &  177.47\\ 
        \bottomrule
    \end{tabular}}
    \end{minipage}
    \begin{minipage}{0.49\textwidth}
    \centering
    \caption{Distribution by Reimbursement Class (AIFA Classification)}
    \label{tab:classe-farmaceutica}
    \scalebox{0.95}{
    \rowcolors{2}{white}{gray!10}
    \begin{tabular}{lc}
        \toprule
        \textbf{Class Code} & \textbf{Frequency} \\
        \midrule
        C     & 5,406 \\
        A     & 5,156 \\
        N     & 1,842 \\
        C-nn  & 1,724 \\
        H     & 1,599 \\
        C-bis & 942  \\
        \bottomrule
    \end{tabular}}
    \end{minipage}
\end{table*}

The reimbursement classes reflect regulatory categories distinct from therapeutic classes (including six ATC Level 1 codes). Classes C and A dominate the dataset, representing non-reimbursed and essential reimbursed drugs, respectively. The subclasses C-nn and C-bis are nested under C, which explains the sum exceeding the total document count.

\paragraph{Section Coverage and Quality Metrics.}
We evaluated the presence of key regulatory sections across documents to assess completeness and usability for NLP tasks. Table~\ref{tab:section_coverage} reports the coverage of selected sections critical for pharmacological information extraction. The results indicate a high degree of consistency, with most sections present in over 90\% of the RCPs, ensuring reliable availability of therapeutic indications, dosage information, contraindications, interactions, and adverse effects for computational analysis. Only section 04.6 Pregnancy/Lactation has a slightly lower percentage of coverage (89.6\%).

\begin{table}[ht]
    \centering
    \caption{Coverage of key sections of the RCPs in the dataset \texttt{DART}}
    \label{tab:section_coverage}
    \rowcolors{2}{gray!10}{white}
    \begin{tabular}{lll}
        \toprule
        \textbf{Section Code} & \textbf{Section Name}       & \textbf{Coverage (\%)} \\
        \midrule
        04.1 & Therapeutic Indications      & 97.4 \\
        04.2 & Posology and Use             & 95.9 \\
        04.3 & Contraindications            & 94.1 \\
        04.5 & Interactions                 & 92.7 \\
        04.6 & Pregnancy/Lactation          & 89.6 \\
        04.8 & Undesirable Effects          & 93.2 \\
        \bottomrule
    \end{tabular}
\end{table}

\paragraph{Lexical and Semantic Insights.}
The vocabulary size ($\approx$103k unique terms) includes a significant portion of technical jargon, multi-token entities, and standard pharmaceutical terminology.
Key pharmacological terms (e.g., “interactions”, “pregnancy”, “contraindications”) occur with high frequency across classes, supporting targeted NLP extraction. 
Further lexical analysis is ongoing to quantify the proportion of domain-specific terms and evaluate term distribution across reimbursement and therapeutic classes.

\section{Applications}

The \texttt{DART} dataset derived from RCPs supports high-precision tasks in computational pharmacovigilance, structured biomedical information extraction, and explainable clinical decision support systems. The dataset applies a semantic structuring pipeline, which categorizes regulatory text into standardized groups (e.g., interactions, contraindications, adverse effects), allowing traceable links to source sections, aligning with regulatory demands, and enhancing interpretability, notably during clinical or legal audits. This structured, authoritative data anchors AI-driven systems, fostering the development of reliable, explainable tools for clinicians, pharmacists, and health IT systems. Key applications of this dataset are outlined below.

\subsection{LLM-based Drug-Drug Interaction Checker}

\begin{figure}
\centering
\includegraphics[width=0.9\linewidth]{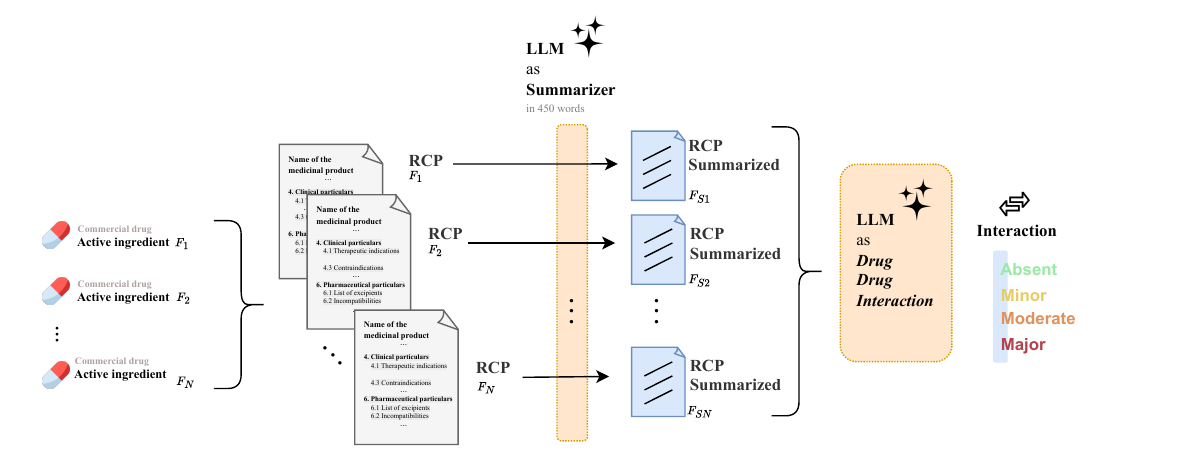}
\caption{System pipeline for automatic detection of drug–drug interactions (DDIs). Each commercial drug, associated with its active ingredient $(F_1, F_2, ..., F_N)$ is linked to its respective Summary of RCPs. From each RCP, the system extracts clinically and pharmacologically relevant sections. These sections are then processed by the Summarizer component, it is an LLM which generates a structured summary (max. 450 words). The resulting summaries are fed into a LLM, which analyzes all drug pairs and classifies the interaction severity as absent, minor, moderate, or major with prompt engineering.}
\label{fig:drugdrug}
\end{figure}

To assess the effectiveness of the \texttt{DART} dataset in the context of automated processing of regulatory information, we designed and implemented an advanced system for the identification of drug–drug interactions (DDIs), leveraging the capabilities of LLMs. The system takes as input a set of drugs $F = {(F_1, F_2, ..., F_N)}$ each represented through its active ingredient and the structured sections of the RCP, extracted directly from the \texttt{DART} dataset.
RCPs, being rich and complex technical documents, contain relevant information for DDI detection dispersed across heterogeneous sections such as “Warnings and Precautions”, “Interactions”, or “Pharmacokinetic Properties”. However, direct analysis of the full text proves suboptimal for LLMs due to both input length limitations and high semantic dispersion. In order to tackle these issues, as outlined in Section~\ref{sec: RCP_summarization}, we implemented a regulatory summarization module in which each drug is paired with a corresponding summary, denoted as $F_S = (F_{S1}, F_{S2}, ... F_{SN})$. This feature, built upon an LLM, produces an organized summary concentrating solely on components that could be pertinent to analyzing pharmacological interactions. In this initial phase, the system markedly reduces the complexity of the regulatory text, directing the model's focus toward clinically relevant concepts while ensuring compliance with the computational constraints of current LLMs. The resulting summary is then forwarded to the LLM-as-DDI module, which—through the application of targeted prompt engineering techniques—detects potential interactions between active pharmaceutical ingredients, elucidates the underlying pharmacological mechanisms (such as receptor synergies or enzymatic pathways), and assesses the clinical relevance of each interaction based on the patient’s profile. This process is followed by the formulation of context-specific recommendations—such as dosage adjustments or monitoring requirements—and the assignment of a severity level to each identified interaction. The full pipeline is illustrated in Figure~\ref{fig:drugdrug} and an example end-to-end is showed in Figure~\ref{fig:ddi_dart_horizontal}.
Interactions are categorized into four ascending levels of clinical severity—\textit{Absent}, \textit{Minor}, \textit{Moderate}, and \textit{Major}—in accordance with taxonomies commonly employed in scientific research. Specifically, \textit{Absent} indicates the lack of any known or clinically meaningful interaction between the drugs; \textit{Minor} denotes a pharmacological interaction of negligible clinical relevance, typically not requiring any intervention; \textit{Moderate} refers to a clinically significant interaction that may necessitate monitoring or dosage adjustments; and \textit{Major} implies a severe interaction, which is either contraindicated or requires substantial modifications to the therapeutic regimen. To facilitate comparison with widely used online tools such as Drugs.com\footnote{\url{https://www.drugs.com/drug_interactions.html}}, Medscape\footnote{\url{https://reference.medscape.com/drug-interactionchecker}}, WebMD\footnote{\url{https://www.webmd.com/interaction-checker/default.htm}}, and RxList\footnote{\url{https://www.rxlist.com/drug-interaction-checker.htm}}, which adopt a binary classification framework, we employed a simplified model that consolidates the \textit{Minor}, \textit{Moderate}, and \textit{Major} categories into a single class, labeled \texttt{Interaction}, while retaining \texttt{Absent} as a distinct category. This adaptation ensures compatibility with systems commonly used in clinical practice.
Performance was evaluated using standard metrics—Precision, Recall, F1-score, and Accuracy—on a manually annotated test set comprising 100 examples. Particular emphasis was placed on Recall, as it serves as a critical metric for assessing the system’s ability to detect all clinically relevant drug–drug interactions (DDIs). In medical contexts, achieving high Recall is essential to minimize false negatives and thereby ensure patient safety.
Table~\ref{tab:ddi-comparison} highlights the substantial advantages of the proposed framework. The upper section of the table presents the results obtained from four established web-based tools. A range of large language models (LLMs) were tested in standalone configuration, including both closed-source models (GPT-4o, Claude, Gemini) and open-source models (LLaMA, Mistral, Gemma). Additionally, the table includes performance data for open-source models enhanced with regulatory summaries generated via the \texttt{DART} system.
Comparative analysis reveals that certain closed-source models, such as Claude-3.5 and GPT-4o, achieve performance comparable to or exceeding that of conventional clinical tools. However, the incorporation of \texttt{DART} summarization emerges as a pivotal factor. For instance, the configuration LLaMA-3.1-8B + \texttt{DART} achieves a Recall of 0.843, substantially outperforming the same model without summarization (Recall = 0.229), and surpassing most of the evaluated web-based systems.
These findings underscore the critical role of guided regulatory summarization in enhancing DDI detection capabilities without compromising precision. Overall, the results validate the effectiveness of the proposed framework: integrating the \texttt{DART} dataset with advanced language models enables even lightweight open-source architectures to effectively identify complex pharmacological interactions. This approach demonstrates the potential to rival high-end proprietary systems, offering an optimal balance of accuracy, coverage, and computational efficiency—factors essential for practical deployment in both clinical and regulatory domains.

\begin{table}[ht]
\caption{Comparative Evaluation of DDI Detection Tools: Web-based Systems, Standalone LLMs (Closed/Open Source), and LLMs Enhanced with the \texttt{DART} Framework. Results are reported for binary classification performance (\texttt{Absent} vs. \texttt{Interaction}).}
\label{tab:ddi-comparison}
\centering
\rowcolors{2}{white}{gray!10}
\begin{tabular}{lcccc}
\toprule
\multicolumn{5}{c}{\textbf{Web-based Drug Interaction Tools}} \\
\toprule
\textbf{Tool} & \textbf{Precision} & \textbf{Recall} & \textbf{F1-score} & \textbf{Accuracy} \\
\midrule
Drugs.com\footnote{\url{https://www.drugs.com/drug_interactions.html}} & 0.701 & 0.812 & 0.756 & 0.76 \\
Medscape\footnote{\url{https://reference.medscape.com/drug-interactionchecker}} & 0.761 & 0.712 & 0.734 & 0.74 \\
WebMD\footnote{\url{https://www.webmd.com/interaction-checker/default.htm}} & 0.682 & 0.673 & 0.675 & 0.675 \\
RxList\footnote{\url{https://www.rxlist.com/drug-interaction-checker.htm}} & 0.663 & 0.633 & 0.645 & 0.645 \\
\midrule

\multicolumn{5}{c}{\textbf{LLM Standalone}} \\
\toprule
\textbf{Model} & \textbf{Precision} & \textbf{Recall} & \textbf{F1-score} & \textbf{Accuracy} \\
\midrule
\multicolumn{5}{l}{\textit{Closed-source Models}} \\
\midrule
Gemini-2.5-Flash & 0.814 & 0.500 & 0.619 & 0.657 \\
GPT-4o & 0.688 & 0.786 & 0.733 & 0.737 \\
Claude-3.5 & 0.730 & 0.771 & 0.750 & 0.750 \\
\midrule
\multicolumn{5}{l}{\textit{Open-source Models}} \\
\midrule
Gemma-2-9b & 0.750 & 0.214 & 0.333 & 0.554 \\
LLaMA-3.1-8B & 0.800 & 0.229 & 0.356 & 0.561 \\
Mistral-7B & \textbf{0.842} & 0.229 & 0.360 & 0.565 \\
\midrule

\multicolumn{5}{c}{\textbf{LLM + \texttt{DART} (Our Demonstration)}} \\
\toprule
\textbf{Model} & \textbf{Precision} & \textbf{Recall} & \textbf{F1-score} & \textbf{Accuracy} \\
\midrule
Gemma-2-9b + \texttt{DART} & 0.711 & 0.457 & 0.557 & 0.684 \\
LLaMA-3.1-8B + \texttt{DART} & 0.728 & \textbf{0.843} & \textbf{0.781} & \textbf{0.786} \\
Mistral-7B + \texttt{DART} & 0.778 & 0.300 & 0.433 & 0.640 \\
\bottomrule
\end{tabular}
\end{table}

\begin{figure}[!t]
\centering
\begin{tcolorbox}[width=0.96\textwidth,
                  colback=gray!5,
                  colframe=blue!20,
                  coltitle=black,
                  title=\textbf{Illustrative Example of Drug–Drug Interaction Detection using the DART Framework},
                  enhanced]

\textbf{Input:} \\
\textit{Drug F1}: Warfarin \hfill \textit{Drug F2}: Ibuprofen \\
Active Ingredient: Warfarin \hfill Active Ingredient: Ibuprofen

\vspace{0.8em}
\textbf{Step 2 – Extract Summarized RCPs:} \\
RCP for Warfarin $\rightarrow$ Summarized RCP F1 ($\approx$ 450 words) \\ RCP for Ibuprofen $\rightarrow$ Summarized RCP F2 ($\approx$ 450 words)

\vspace{0.8em}
\textbf{Step 3 – Compare Summaries to Detect Interaction:} \\
The LLM receives the summarized RCPs F1 and F2, then prompting them $\rightarrow$ Evaluates interaction risk, mechanism, and severity

\vspace{0.8em}
\textbf{Output:} \\
Interaction Detected: \textbf{Major} \\
Drug Pair: Warfarin + Ibuprofen \\
Mechanism: Inhibition of CYP2C9 by ibuprofen increases the bleeding risk associated with warfarin \\
Recommendation: Avoid co-administration or closely monitor INR levels

\end{tcolorbox}
\caption{An end-to-end example of drug–drug interaction detection using the \texttt{DART} framework. The system retrieves, summarizes, and compares  documents to assess clinical relevance.}
\label{fig:ddi_dart_horizontal}
\end{figure}

\subsection{Other Applications}

\paragraph{Training and Fine-tuning of Multilingual NLP Models.}
The dataset \texttt{DART} serves as a natural benchmark for training and fine-tuning NLP models specialized in Named Entity Recognition (NER) and Relation Extraction (RE) in Italian, particularly for regulatory and clinical pharmacology domains. It includes entities such as active substances, administration routes, pharmacokinetic mechanisms, and pregnancy risk categories. Thanks to its semantic consistency and structural regularity, the dataset supports both supervised training and distant supervision, filling a critical gap in the multilingual biomedical NLP landscape, which remains heavily English-centric.

\paragraph{Fine-tuning of Domain-specific LLMs or SLMs.}
The normalized corpus of RCP texts offers a unique foundation for domain-specific fine-tuning of LLMs or Small Language Models (SLMs) tailored to the Italian pharmaceutical regulatory domain. Potential downstream applications include: Automatic classification of clinical risks from free text; Assisted generation of pharmacovigilance reports; Controlled rewriting of regulatory documents (e.g., technical leaflets, RCPs).
Such models could significantly enhance automation and consistency in regulatory workflows, particularly in contexts requiring traceable and explainable outputs.

\paragraph{Construction of Regulatory Knowledge Graphs.}
The extracted relational triples (e.g., active substance $\rightarrow$ causes $\rightarrow$ adverse effect, drug $\rightarrow$ interacts with $\rightarrow$ compound) can be transformed into semantic knowledge graphs (KGs). These KGs support automated inference over contraindications and interactions, and allow structured linking between regulatory sources (e.g., RCPs) and observational data (e.g., national registries such as RNF or VALORE). Moreover, KGs facilitate the generation of explainable clinical decision rules, increasing transparency and trust in AI-powered systems.

\paragraph{Semi-automated Population of Clinical Decision Support Systems (CDSS).}
The structured dataset is highly suitable for integration into next-generation Clinical Decision Support Systems that combine structured knowledge (e.g., ontologies, terminologies) with unstructured textual evidence. Data extracted from RCPs can be used to populate modules within Electronic Health Records (EHRs), generate safety alerts in hospital pharmacy systems, or support real-time prescription checks. The goal is to enhance patient safety and prescribing appropriateness, by anchoring decisions to verified, regulatory-grade information.

\section{Conclusion \& Future Work}

This work introduces a structured and scalable method for transforming Italian Summary of Product Characteristics (RCPs) into machine-readable resources for biomedical AI. Through semantic parsing and organization, we demonstrate their applicability in multiple domains, including pharmacological interaction checking, domain-specific model tuning, knowledge graph creation, and clinical decision support systems. Despite their linguistic variability, RCPs offer a strong foundation for transparent and regulation-compliant AI systems. The resulting dataset serves both as a benchmark for multilingual biomedical NLP and as a driver for innovation in pharmacovigilance and clinical AI. Future developments will aim to extend coverage to additional regulatory document types and therapeutic areas, improve prompt and alignment strategies, introduce validation processes with domain experts, and publish reusable tools and subsets to support open research in regulatory science.

\paragraph{Limitations}
Although \texttt{DART} represents a relevant step for Italian biomedical NLP, some limitations apply. It includes only RCPs, thus lacking real-world clinical nuances such as patient adherence or off-label use. Not all AIFA-listed medicines are included due to technical issues like malformed or inaccessible documents, potentially underrepresenting some drug categories. Additionally, the LLM-based components, while optimized for factual consistency, may miss rare or context-specific details and remain sensitive to prompt design and model variability.

\paragraph{Ethical Issues}
Since \texttt{DART} relies exclusively on publicly available regulatory texts intended for healthcare professionals, it presents minimal direct ethical risks. However, caution is necessary when using generated outputs in clinical contexts, as language models may propagate inaccuracies, particularly in sensitive areas like drug safety. Human oversight remains essential, and future work should include expert review and mechanisms to flag uncertainty.

\paragraph{Data License and Copyright Issues}
All documents were sourced from the official website of the Italian Medicines Agency (AIFA) under public access policies. The \texttt{DART} dataset is released under the \textit{Creative Commons Attribution 4.0 International License (CC BY 4.0)}, allowing reuse with proper attribution. However, the original RCPs remain property of AIFA, and downstream use must respect ethical and regulatory guidelines.

\begin{acknowledgments}
   This work was conducted with the financial support of (1) the PNRR MUR project PE0000013-FAIR and (2) the Italian ministry of economic development, via the ICARUS (Intelligent Contract Automation for Rethinking User Services) project (CUP: B69J23000270005).
\end{acknowledgments}

\section*{Declaration on Generative AI}
During the preparation of this work, the author(s) used ChatGPT and DeepL in order to: Grammar and spelling check. After using these tool(s)/service(s), the author(s) reviewed and edited the content as needed and take(s) full responsibility for the publication’s content.

\bibliography{sample-ceur}




\end{document}